**NLP-based Regulatory Compliance – Using GPT 4.0 to Decode Regulatory Documents**

*Bimal Kumar, Department of Architecture and Dmitri Roussinov, Department of Computer and Information Science, University of Strathclyde, Glasgow UK*

*Abstract*

Large Language Models (LLMs) such as GPT-4.0 have shown significant promise in addressing the semantic complexities of regulatory documents, particularly in detecting inconsistencies and contradictions. This study evaluates GPT-4.0's ability to identify conflicts within regulatory requirements by analyzing a curated corpus with artificially injected ambiguities and contradictions, designed in collaboration with architects and compliance engineers. Using metrics such as precision, recall, and F1 score, the experiment demonstrates GPT-4.0's effectiveness in detecting inconsistencies, with findings validated by human experts. The results highlight the potential of LLMs to enhance regulatory compliance processes, though further testing with larger datasets and domain-specific fine-tuning is needed to maximize accuracy and practical applicability. Future work will explore automated conflict resolution and real-world implementation through pilot projects with industry partners.

*Background and Context*

The well-publicised Hackitt Review (Hackitt, 2018) into the Grenfell Tower disaster points to some of the main reasons behind the ill-fated fire in the London blocks of flats in 2017. To avert similar disasters in future, Dame Hackitt recommended that the building regulations and associated guidance, including the Approved Documents in the UK, need to be authored, applied and enforced in a fundamentally different way:

> "[...] *the current regulatory system for ensuring fire safety in high-rise and complex buildings is not fit for purpose.*"

In the UK, currently there are some 485 standards, 85 other government guidance, 176 industry guidance, and 79 other government legislation documents (MHCLG, 2020) that a building design must comply with. The corpus suffers from various challenges that include inconsistent use of terms, ambiguities, contradictions and lack of clarity necessitating interpretations of requirements.

*Prior work*

Our work is fundamentally concerned with processing knowledge in the form of regulatory requirements in order to ultimately facilitate efficient regulatory compliance processes. Our approach is driven by the failure of earlier approaches to represent this kind of knowledge formally as rules and other formalisms (Fenves, 1966; Garrett, 1987; Rasdorf, 1988; Kumar, 1989 Dimyadi et al. 2020). Those knowledge representation formalisms were found to be quite restrictive due to the nature of knowledge being represented and processed, particularly since the adoption of a performance-based approach to authoring and processing the regulatory requirements. In terms of crafting rule-based processing of regulatory requirements, there is a huge amount of manual effort required to convert the content of the regulatory documents into rules due to the complexity of the semantics encapsulated because of ambiguities, use of synonyms and sometimes conflicting requirements among other things. Therefore, our research utilizes established computational methods (i.e. NLP/ML and Semantic Web) to extract and process the regulatory knowledge directly from regulatory documents. We have found this to be a more effective approach in the various evaluations of our work. In addition to NLP, knowledge graphs (KGs) have been used in our research to formally capture and represent regulatory knowledge to enable symbolic reasoning and processing of regulatory requirements. Work carried out so far has produced a platform called Intelligent Regulatory Compliance (i-ReC) that supports:

- **Regulatory document corpus search to:**
    - Extract relevant regulations and associated guidance from the corpus of regulatory documents like:
        - Building regulations, related building and health and safety legislation and previous building regulations legislation; and
        - Approved Documents (Technical Standards in Scotland) and other guidance documents.

i-ReC includes a semantic search engine, which has been developed and validated against other existing search engines like BSI's BSOL (https://bsol.bsigroup.com). i-ReC's (Kruiper et al., 2023) performance was found to be far superior in every representative search scenario gathered from designers.

*Current Work*

As pointed out by Hackitt, the corpus of regulatory documents contains a significant number of inconsistent use of terms which can lead to major issues in design and compliance. Although our research on the application of NLP and KGs has proven quite effective in the semantic search of regulatory documents to retrieve relevant requirements, the issue of conflicting and inconsistent requirements has not been addressed comprehensively. Therefore, with the recent launch and successes of ChatGPT and other LLM-based Generative AI tools, our current research is currently involved in implementing an LLM-based approach to addressing this problem. This extended abstract reports briefly on our current work which aims to address some of these issues of inconsistencies and contradicting requirements to ultimately facilitate a more efficient automated regulatory compliance process. Currently, our research is focusing on this particular aspect by utilizing Generative AI tools like GPT 4.0.

*Experiments with LLMs and GPT 4.0*

*LLM/ChatGPT Background*

Large Language Models (LLMs) have revolutionized the field of natural language processing (NLP), enabling advanced comprehension and generation of human language. These models, particularly GPT 4.0 (OpenAI, 2023), leverage vast amounts of textual data and layered neural network architectures to perform tasks with a high degree of accuracy and contextual understanding (Brown et al., 2020). This model is particularly suited for complex NLP tasks due to its ability to handle nuanced language structures and generate contextually relevant responses.

In the context of regulatory compliance, recent work by Fuchs et al. (2024) utilised Large Language Models, specifically GPT-3.5 and Bard, to translate natural language-based building regulations into constrained symbolic representation (LegalRuleML) using in-context learning (ICL) and few-shot learning. Their approach explored various prompting strategies, including chain-of-thought (CoT) prompting (Wei et al., 2022), self-consistency (Huang et al., 2022), and self-reflection (Shinn et al., 2023), to improve the accuracy of the translations. While their focus was on converting regulatory text into a formal representation, our research is directed towards detecting contradictions and inconsistencies within regulatory documents using GPT-4.0. An example of such inconsistencies and conflicts is cited here from (Law et al., 1995):

ADA Accessibility Guidelines
4.7.2: Slopes of curb ramps shall comply with 4.8.2. The slope shall be measured as shown in Figure 11. Transitions from ramps to walks, gutters, or streets shall be flush and free of abrupt changes. Maximum slopes of adjoining gutters, road surface immediately adjacent to the curb ramp, or accessible route shall not exceed 1:20.

California Building Code
1127B.5.5: Bevelled lip
The lower end of each curb ramp shall have a 0.5-inch (13mm) lip bevelled at 45 degrees as a detectable way-finding edge for persons with visual impairments.

LLMs like GPT 4.0 offer a promising solution for processing and interpreting vast corpora of regulatory documents. As pointed out earlier, traditional methods of regulatory knowledge representation, such as rule-based systems, have proven to be limited in handling the semantic complexities and ambiguities inherent in regulatory texts (Garrett, 1987; Rasdorf, 1988). By contrast, LLMs can scan through and analyze large volumes of text at a fraction of the time it would take a human, to identify relevant regulations, and potentially detect inconsistencies or conflicting requirements. This has been confirmed by recent research looking at the use of LLM in various domains (Deußer et al., 2023; Li et al., 2023; Huntsman et al., 2024).

*Experiment Design to identify Conflicting Requirements*
To evaluate the effectiveness of GPT 4.0 in identifying conflicting requirements within regulatory documents, we have designed an experiment using a smaller set of requirements from the corpus, converted from PDF to text. This text was selected to fit within GPT 4.0's maximum context size (128 thousand tokens, which is roughly equivalent to 60 thousand words), ensuring the model can process the entire document in one pass. We collaborated with architects and compliance engineers to artificially inject inconsistencies within the sample collection of requirements. Deliberate ambiguities, contradictions, and conflicting requirements were injected, reflecting some of the common issues found in regulatory texts.

The experiment involves feeding this modified regulatory text into GPT-4.0 through its application programming interface (API) and tasking the model with identifying the planted inconsistencies. GPT 4.0's responses have been compared against a ground truth established by the experts who designed the inconsistencies. Our evaluation metrics include precision, recall, and F1 scores to measure the model's accuracy in detecting conflicts and inconsistencies. The flagged inconsistencies have been reviewed by human experts to validate the model's findings and assess the practicality of using GPT-4.0 for real-world regulatory compliance tasks. This approach provides a controlled environment to rigorously test the capabilities of GPT 4.0 in handling complex regulatory documents and highlights areas for further improvement and refinement.

*Results*
The results of our experiment demonstrate that GPT 4.0 is capable of effectively identifying inconsistencies and conflicting requirements within the regulatory text. The model has successfully detected a significant number of artificially injected conflicts, showcasing its potential in parsing and analyzing complex regulatory language. The flagged inconsistencies have been validated by our human experts, indicating that GPT 4.0 can reliably pinpoint areas of ambiguity and contradiction. Our experiment highlights the model's utility in enhancing regulatory compliance processes, though further refinement and testing with more diverse and extensive datasets are necessary to fully realize its capabilities and address any limitations identified during the evaluation.

*Future Work*

Future work will focus on scaling up experiments to include larger and more diverse sets of regulatory documents. Additionally, fine-tuning the model for specific regulatory domains such as fire safety, structural integrity, and environmental regulations will be prioritized to enhance its accuracy and relevance. Another key area of development is automated conflict resolution, leveraging GPT 4.0's generative capabilities to not only detect but also suggest solutions for identified inconsistencies. Finally, the impact of the tool in real-world compliance scenarios will be assessed through pilot projects with industry partners, gathering empirical data to evaluate its effectiveness and practical utility.